\documentclass[11pt]{article}
\usepackage{acl2016}
\usepackage{times}
\usepackage{url}
\usepackage{latexsym}

\usepackage{array}
\usepackage{amsopn}
\usepackage{xcolor}
\usepackage{graphicx}
\usepackage{multirow}
\usepackage{amssymb}
\usepackage{amsmath}
\usepackage[font={footnotesize}]{caption}

\newcommand{\Reals}{\mathbb{R}}

\author{
Daniel Hewlett, Alexandre Lacoste, Llion Jones, Illia Polosukhin, \\
\textbf{Andrew Fandrianto, Jay Han, Matthew Kelcey and David Berthelot }\\ Google Research \\ \small{{\tt \{dhewlett,allac,llion,ipolosukhin,fto,hanjay,matkelcey,dberth\}@google.com }} }
  
\aclfinalcopy % Uncomment this line for the final submission
\def\aclpaperid{563} %  Enter the acl Paper ID here

\DeclareMathOperator{\softmax}{softmax}

\title{\textsc{WikiReading}: A Novel Large-scale Language Understanding Task over Wikipedia}
 
\date{}
 
\begin{document}
 
\maketitle 

%TODO from joberant:
%Here are my comment for the abstract (we can also take vis-a-vis, probably that would be good): 
%- This is something I wrote already (and it’s fine if you choose to ignore it), but I think it is good to start with statement of the problem you are trying to solve, something about language understanding being very interesting, but there are not enough large scale dataset that allow end to end learning or something (but concisely!). So I would present the problem and then say, “in this paper we present WIkiReading…. 
%To conclude - I think the abstract needs tightening. Make sure there is a clear flow and things are phrased very precisely.

\begin{abstract}

We present \textsc{WikiReading}, a large-scale natural language understanding task and publicly-available dataset with 18 million instances. 
The task is to predict textual values from the structured knowledge base Wikidata by reading the text of the corresponding Wikipedia articles.
The task contains a rich variety of challenging classification and extraction sub-tasks, making it well-suited for end-to-end models such as deep neural networks (DNNs).
%, presenting a challenge for current state of the art algorithms. 
%To enable , we collected a benchmark dataset containing 18 million instances. 
%We evaluate a number of recent DNN-based architectures for document classification, information extraction, and question answering on the \textsc{WikiReading} benchmark.
%We find that sequence to sequence learning possesses fewer intrinsic limitations than other approaches and outperforms them, especially with appropriate pre-training.
%In particular, the best-performing model treats both the input and output as raw byte sequences.
We compare various state-of-the-art DNN-based architectures for document classification, information extraction, and question answering.
We find that models supporting a rich answer space, such as word or character sequences, perform best. 
Our best-performing model, a word-level sequence to sequence model with a mechanism to copy out-of-vocabulary words, obtains an accuracy of 71.8\%.

\end{abstract}

\section{Introduction}\label{sec:intro}

A growing amount of research in natural language understanding (NLU) explores end-to-end deep neural network (DNN) architectures for tasks such as text classification \cite{zhang2015character}, relation extraction \cite{nguyen2015relation}, and question answering \cite{DBLP:journals/corr/WestonBCM15}.
These models offer the potential to remove the intermediate steps traditionally involved in processing natural language data by operating on increasingly raw forms of text input, even unprocessed character or byte sequences.
%This may allow for better overall performance by tuning lower-level processing towards the final objective, and makes extension to new languages easier.
Furthermore, while these tasks are often studied in isolation, DNNs have the potential to combine multiple forms of reasoning within a single model. 

Supervised training of DNNs often requires a large amount of high-quality training data. 
To this end, we introduce a novel prediction task and accompanying large-scale dataset with a range of sub-tasks combining text classification and information extraction. The dataset is made publicly-available at \url{http://goo.gl/wikireading}.
The task, which we call \textsc{WikiReading}, is to predict textual values from the open knowledge base \emph{Wikidata} \cite{VK:wikidata14} given text from the corresponding articles on \emph{Wikipedia} \cite{ayers2008wikipedia}. 
Example instances are shown in Table~\ref{tab:examples}, illustrating the variety of subject matter and sub-tasks.
The dataset contains 18.87M instances across 867 sub-tasks, split roughly evenly between classification and extraction (see Section \ref{sec:wikireading} for more details).

% Total instances: 18,867,478
% Total properties: 867

\begin{table*}
\small
\begin{tabular}{|p{1.4cm}|p{3.1cm}|p{3.1cm}|p{3.1cm}|p{3.1cm}|}
\hline
& \multicolumn{2}{c}{{\bf Categorization}} \vline & \multicolumn{2}{c}{{\bf Extraction}} \vline \\
\hline
{\bf Document} & 
Folkart Towers are twin skyscrapers in the Bayrakli district of the Turkish city of Izmir. Reaching a structural height of 200 m (656 ft) above ground level, they are the tallest \dots &
Angeles blancos is a Mexican telenovela produced by Carlos Sotomayor for Televisa in 1990. Jacqueline Andere, Rogelio Guerra and Alfonso Iturralde star as the main \dots &
% {\bf Rubus} deamii, Deam's dewberry, is an uncommon North American species of flowering plant in the rose family. It grows in scattered locations in the east-central United States\dots &
Canada is a country in the northern part of North America. Its ten provinces and three territories extend from the {\bf Atlantic} to the {\bf Pacific} and northward into the {\bf Arctic Ocean}, \dots &
%FC Palmira Odesa was a professional football club based in Odesa, Ukraine. The club was founded in {\bf 1999} under the name Chornomorets-Lasunya\dots 
Breaking Bad is an American crime drama television series created and produced by Vince Gilligan. The show originally aired on the AMC network for five seasons, from {\bf January 20, 2008}, to \dots 
\\\hline
{\bf Property} & country & original language of work & located next to body of water & start time \\\hline\hline
{\bf Answer} & Turkey & Spanish & Atlantic Ocean, Arctic Ocean, Pacific Ocean & 20 January 2008 \\\hline
\end{tabular}
\caption{Examples instances from \textsc{WikiReading}. The task is to predict the answer given the document and property. Answer tokens that can be extracted are shown in bold, the remaining instances require classification or another form of inference.}\label{tab:examples}
\end{table*}

In addition to its diversity, the \textsc{WikiReading} dataset is also at least an order of magnitude larger than related NLU datasets.
%than the news dataset from Hermann et al. \shortcite{DBLP:journals/corr/HermannKGEKSB15}. 
Many natural language datasets for question answering (QA), such as \textsc{WikiQA} \cite{yang2015wikiqa}, have only thousands of examples and are thus too small for training end-to-end models.
\newcite{DBLP:journals/corr/HermannKGEKSB15} proposed a task similar to QA, predicting entities in news summaries from the text of the original news articles, and generated a \textsc{News} dataset with 1M instances.
%\textsc{WikiQA} and \textsc{News} require only pointing to locations within the document.
The bAbI dataset \cite{DBLP:journals/corr/WestonBCM15} requires multiple forms of reasoning, but is composed of synthetically generated documents.
\textsc{WikiQA} and \textsc{News} only involve pointing to locations within the document, and text classification datasets often have small numbers of output classes. In contrast, \textsc{WikiReading} has a rich output space of millions of answers, making it a challenging benchmark for state-of-the-art DNN architectures for QA or text classification.

%TODO from joberant
%transition from paragraph 3 to 4 is a bit weird. 
%Right now 3 is about comparison to other datasets. 
%4 is also about that but also about models. 
%You should perhaps have 3 only compare to other datasets, and 4 only talk about your methods. 

We implemented a large suite of recent models, and for the first time evaluate them on common grounds, placing the complexity of the task in context and illustrating the tradeoffs inherent in each approach.
The highest score of 71.8\% is achieved by a sequence to sequence model \cite{DBLP:journals/corr/KalchbrennerB13,cho-al-emnlp14} operating on word-level input and output sequences, with special handing for out-of-vocabulary words.
%These promising results suggest that \textsc{WikiReading} provides 
%We show that careful training of these models is essential, as a pre-training procedure

\section{\textsc{WikiReading}}\label{sec:wikireading}

%TODO from joberant:
%One more general comment about Section 2. 
%One thing that will make this paper accepted is if people believe that this dataset is challengin an interesting. I think Section 2 is a good place to illustrate that using examples. Perhaps you should have a paragraph/subsection on that. Specifically have some text that argues why this is a really cool task. Probably if people are convinced of that the paper will be accepted. Right now there are not enough examples (I suggested a table in the beginning, but that's not enough) to get a sense and numbers only are also not enough. 
%
%In general I think examples, tables and visualizations are really good, so I find that that is my main comment for many many papers....

We now provide background information relating to Wikidata, followed by a detailed description of the \textsc{WikiReading} prediction task and dataset.

\subsection{Wikidata}\label{sec:wikidata}
 
Wikidata is a free collaborative knowledge base containing information about approximately 16M \textit{items} \cite{VK:wikidata14}.
Knowledge related to each item is expressed in a set of \textit{statements}, each consisting of a (\texttt{property, value}) tuple. 
For example, the item \texttt{Paris} might have associated statements asserting \texttt{(instance of, city)} or \texttt{(country, France)}. 
Wikidata contains over 80M such statements across over 800 properties.
Items may be linked to articles on Wikipedia.

\subsection{Dataset}\label{sec:dataset}

We constructed the \textsc{WikiReading} dataset from Wikidata and Wikipedia as follows:
We consolidated all Wikidata statements with the same item and property into a single \texttt{(item, property, answer)} triple, where \texttt{answer} is a set of values.
Replacing each \texttt{item} with the text of the linked Wikipedia article (discarding unlinked items) yields a dataset of 18.58M \texttt{(document, property, answer)} instances. 
%TODO from joberant: not a big deal but you say in line 163 "entirely textual", but some might claim that the property is not really language. 
Importantly, all elements in each instance are human-readable strings, making the task entirely textual. The only modification we made to these strings was to convert timestamps into a human-readable format (e.g., ``4 July 1776'').

The \textsc{WikiReading} task, then, is to predict the answer string for each tuple given the document and property strings. 
This setup can be seen as similar to information extraction, or question answering where the property acts as a ``question''.
We assigned each instance randomly to either training (around 16.03M instances), validation (1.89M), and test (0.95M) sets following a 85/10/5 distribution.

\subsection{Documents}

%TODO from joberant: I wonder if the stats will be better expressed in a simple table. 

%\begin{table}
%\small
%
%\begin{tabular}{|l|r|}
%\hline
%Total instances & 18.58M \\\hline
%Unique articles & 4.7M \\\hline
%Unique properties & 884 \\\hline
%Unique answers & \emph{8M} \\\hline
%Avg instances per article & 5.31 \\\hline
%Avg answers per article
%\end{tabular}
%
%\caption{Dataset statistics.}\label{tab:dataset-stats}
%\end{table}
%

The dataset contains 4.7M unique Wikipedia articles, meaning that roughly 80\% of the English-language Wikipedia is represented. Multiple instances can share the same document, with a mean of 5.31 instances per article (median: 4, max: 879). The most common categories of documents are \texttt{human}, \texttt{taxon}, \texttt{film}, \texttt{album}, and \texttt{human settlement}, making up 48.8\% of the documents and 9.1\% of the instances. The mean and median document lengths are 489.2 and 203 words. 

% TODO: These stats may be from the old dataset.

%In Wikipedia articles, the opening paragraph is intended to summarize the essential information about the entity being described, and thus a disproportionate number of answers can be found in or deduced from the early parts of the documents. When one of the values within an answer appears verbatim in the instance's document, it occurs within the first 300 words X\% of the time. 

\subsection{Properties}\label{sec:properties}

The dataset contains 867 unique properties, though the distribution of properties across instances is highly skewed:
% TODO: Maybe move to table?
The top 20 properties cover 75\% of the dataset, with 99\% coverage achieved after 180 properties.
We divide the properties broadly into two groups: 
\textit{Categorical} properties, such as \texttt{instance of}, \texttt{gender} and \texttt{country}, require selecting between a relatively small number of possible answers, while \textit{relational} properties, such as \texttt{date of birth}, \texttt{parent}, and \texttt{capital}, typically require extracting rare or totally unique answers from the document.

To quantify this difference, we compute the entropy of the answer distribution $A$ for each property $p$, scaled to the $[0, 1]$ range by dividing by the entropy of a uniform distribution with the same number of values, i.e., $\hat{H}(p) = H(A_p) / \log{|A_p|}$. 
Properties that represent essentially one-to-one mappings score near $1.0$, while a property with just a single answer would score $0.0$. 
Table \ref{tab:entropy} lists entropy values for a subset of properties, showing that the dataset contains a spectrum of sub-tasks. 
We label properties with an entropy less than 0.7 as categorical, and those with a higher entropy as relational. 
Categorical properties cover 56.7\% of the instances in the dataset, with the remaining 43.3\% being relational.

% TODO: Update table with final numbers.
% \begin{table}
% \small
% \begin{tabular}{|l|r|c|}
% \hline
% {\bf Property} & {\bf Frequency} & {\bf Entropy}\\
% \hline
% instance of & 2,574,038 & 0.431 \\
% sex or gender & 941,200 & 0.189 \\
% country & 803,252 & 0.536 \\
% date of birth & 785,049 & 0.936 \\
% given name & 767,916 & 0.763 \\
% occupation & 716,176 & 0.589 \\
% country of citizenship & 674,560 & 0.501 \\
% located in \dots entity & 478,372 & 0.802 \\
% place of birth	& 384,951 & 0.800 \\
% date of death & 364,910 & 0.943 \\
% \hline
% \end{tabular}
% \caption{Training set frequency and scaled answer entropy for the 10 most frequent properties.}\label{tab:entropy}
% \end{table}

% 
%  TOTAL INSTANCE IN TRAINING: 16,039,400
\begin{table}
\small
\begin{tabular}{|l|r|c|}
\hline
{\bf Property} & {\bf Frequency} & {\bf Entropy}\\
\hline
instance of             & 3,245,856 & 0.429 \\
sex or gender           & 1,143,126 & 0.186 \\
country                 &   986,587 & 0.518 \\
date of birth           &   953,481 & 0.916 \\
given name              &   932,147 & 0.755 \\ 
occupation              &   869,333 & 0.588 \\
country of citizenship  &   819,301 & 0.508 \\
located in \dots entity &   582,110 & 0.800 \\
place of birth	        &   467,066 & 0.795 \\
date of death           &   442,514 & 0.922 \\
\hline
\end{tabular}
\caption{Training set frequency and scaled answer entropy for the 10 most frequent properties.}\label{tab:entropy}
\end{table}

\subsection{Answers}

The distribution of properties described above has implications for the answer distribution. There are a relatively small number of very high frequency ``head'' answers, mostly for categorical properties, and a vast number of very low frequency ``tail'' answers, such as names and dates. At the extremes, the most frequent answer \texttt{human} accounts for almost 7\% of the dataset, while 54.7\% of the answers in the dataset are unique. There are some special categories of answers which are systematically related, in particular \emph{dates}, which comprise 8.9\% of the dataset (with 7.2\% being unique). This distribution means that methods focused on either head or tail answers can each perform moderately well, but only a method that handles both types of answers can achieve maximum performance. Another consequence of the long tail of answers is that many (30.0\%) of the answers in the test set never appear in the training set, meaning they must be read out of the document. An answer is present verbatim in the document for 45.6\% of the instances.

\section{Methods}\label{sec:methods}

Recently, neural network architectures for NLU have been shown to meet or exceed the performance of traditional methods \cite{zhang2015character,dai2015semi}. 
The move to deep neural networks also allows for new ways of combining the property and document, inspired by recent research in the field of question answering (with the property serving as a question). 
In sequential models such as Recurrent Neural Networks (RNNs), the question could be prepended to the document, allowing the model to ``read'' the document differently for each question \cite{DBLP:journals/corr/HermannKGEKSB15}. 
Alternatively, the question could be used to compute a form of \emph{attention} \cite{bahdanau2014neural} over the document, to effectively focus the model on the most predictive words or phrases \cite{DBLP:journals/corr/SukhbaatarSWF15,DBLP:journals/corr/HermannKGEKSB15}. 
As this is currently an ongoing field of research, we implemented a range of recent models and for the first time compare them on common grounds.
%To provide a sense of the difficulty of this novel task, as well as guide future work in this area, we implemented and evaluated a number of models previously described in the literature. 
We now describe these methods, grouping them into broad categories by general approach and noting necessary modifications. Later, we introduce some novel variations of these models.

\subsection{Answer Classification}\label{sec:answer-classification}

Perhaps the most straightforward approach to \textsc{WikiReading} is to consider it as a special case of document classification. 
%In text classification, the goal is to predict a distribution over class labels given some features derived from the text. 
To fit \textsc{WikiReading} into this framework, we consider each possible answer as a class label, and incorporate features based on the property so that the model can make different predictions for the same document. 
While the number of potential answers is too large to be practical (and unbounded in principle), a substantial portion of the dataset can be covered by a model with a tractable number of answers.
%In the experiments below we used the 50,000 most frequent answers as output classes, which covers 80\% of the instances in the dataset.
%This is possible because, as discussed in Section \ref{sec:dataset}, categorical properties are more frequent than relational ones, and the distribution of answers for each property is often skewed as well.

\subsubsection{Baseline}\label{sec:baseline}

\newcommand{\Nword}{N_{\operatorname{w}}}

The most common approach to document classification is to fit a linear model (e.g., Logistic Regression) over bag of words (BoW) features.
To serve as a baseline for our task, the linear model needs to make different predictions for the same Wikipedia article depending on the property. We enable this behavior by computing two $\Nword$ element BoW vectors, one each for the document and property, and concatenating them into a single $2\Nword$ feature vector.

\subsubsection{Neural Network Methods}

\newcommand{\din}{d_{\operatorname{in}}}
\newcommand{\dout}{d_{\operatorname{out}}}
\newcommand{\Nans}{N_{\operatorname{ans}}}
\newcommand{\Ndoc}{N_{\operatorname{doc}}}
\newcommand{\ans}{\mathbf{y}}
\newcommand{\xb}{\mathbf{x}}
\newcommand{\zb}{\mathbf{z}}

All of the methods described in this section encode the property and document into a \emph{joint representation} $\ans \in \Reals^{\dout}$, which serves as input for a final softmax layer computing a probability distribution over the top $\Nans$ answers. 
Namely, for each answer $i \in \{1,\dots,\Nans\}$, we have:
\begin{equation}
P(i|\mathbf{x}) =  e^{\ans^\top\mathbf{a}_i} / \textstyle\sum_{j=1}^{\Nans}e^{\ans^\top\mathbf{a}_j},
\end{equation}
where $\mathbf{a}_i \in \Reals^{\dout}$ corresponds to a learned vector associated with answer $i$. 
Thus, these models differ primarily in how they combine the property and document to produce the joint representation. 
For existing models from the literature, we provide a brief description and note any important differences in our implementation, but refer the reader to the original papers for further details. 

Except for character-level models, documents and properties are tokenized into words. The $\Nword$ most frequent words are mapped to a vector in $\Reals^{\din}$ using a learned embedding matrix\footnote{Limited experimentation with initialization from publicly-available word2vec embeddings \cite{mikolov2013distributed} yielded no improvement in performance.}. 
Other words are all mapped to a special out of vocabulary (OOV) token, which also has a learned embedding. 
$\din$ and $\dout$ are hyperparameters for these models.

\paragraph{Averaged Embeddings (BoW):} This is the neural network version of the baseline method described in Section \ref{sec:baseline}. 
Embeddings for words in the document and property are separately averaged. The concatenation of the resulting vectors forms the joint representation of size $2\din$.
\paragraph{Paragraph Vector:} We explore a variant of the previous model where the document is encoded as a \emph{paragraph vector} \cite{DBLP:journals/corr/LeM14}. 
We apply the PV-DBOW variant that learns an embedding for a document by optimizing the prediction of its constituent words. 
These unsupervised document embeddings are treated as a fixed input to the supervised classifier, with no fine-tuning.

\paragraph{LSTM Reader:}
This model is a simplified version of the Deep LSTM Reader proposed by \newcite{DBLP:journals/corr/HermannKGEKSB15}. 
In this model, an LSTM \cite{hochreiter1997long} reads the property and document sequences word-by-word and the final state is used as the joint representation. 
This is the simplest model that respects the order of the words in the document. In our implementation we use a single layer instead of two and a larger hidden size. 
More details on the architecture can be found in Section~\ref{sec:training} and in Table~\ref{tab:params}.

\newcommand{\ub}{\mathbf{u}}
\newcommand{\mb}{\mathbf{m}}
\newcommand{\vb}{\mathbf{v}}
\newcommand{\concat}{\operatorname{concat}}
\paragraph{Attentive Reader:} 
This model, also presented in \newcite{DBLP:journals/corr/HermannKGEKSB15}, uses an attention mechanism to better focus on the relevant part of the document for a given property. Specifically, Attentive Reader first generates a representation $\ub$ of the property using the final state of an LSTM while a second LSTM is used to read the document and generate a representation $\zb_t$ for each word. 
Then, conditioned on the property encoding $\ub$, a normalized attention is computed over the document to produce a weighted average of the word representations $\zb_t$, which is then used to generate the joint representation $\ans$. More precisely: 
% \vspace{-5mm}
\begin{align*}
 \mb_t &= \tanh( W_1 \concat(\zb_t,\ub)) \\
 \alpha_t  & = \operatorname{exp} \left( \vb^{\intercal} \mb_t \right) \\
 \mathbf{r} &= \textstyle\sum_t \tfrac{\alpha_t }{\sum_{\tau} \alpha_{\tau}}\zb_t   \\
 \ans &= \tanh( W_2 \concat(\mathbf{r},\ub) ),
\end{align*}
where $W_1$, $W_2$, and $\vb$ are learned parameters.

\paragraph{Memory Network:}
Our implementation closely follows the End-to-End Memory Network proposed in  \newcite{DBLP:journals/corr/SukhbaatarSWF15}.
This model maps a property $p$ and a list of sentences $\xb_1,\dots,\xb_n$ to a joint representation $\mathbf{y}$ by attending over sentences in the document as follows: 
The input encoder $I$ converts a sequence of words $\xb_i = ( x_{i1}, \dots, x_{iL_i} )$ into a vector using an embedding matrix (equation~\ref{eq:memnet_encoder}), where $L_i$ is the length of sentence $i$.\footnote{Our final results use the position encoding method proposed by \newcite{DBLP:journals/corr/SukhbaatarSWF15}, which incorporates positional information in addition to word embeddings.}
The property is encoded with the embedding matrix $U$ (eqn.~\ref{eq:memnet_question}).
Each sentence is encoded into two vectors, a memory vector (eqn.~\ref{eq:memnet_memory}) and an output vector (eqn.~\ref{eq:memnet_output}), with embedding matrices $M$ and $C$, respectively.
The property encoding is used to compute a normalized attention vector over the memories (eqn.~\ref{eq:memnet_attention}).\footnote{Instead of the linearization method of \newcite{DBLP:journals/corr/SukhbaatarSWF15}, we applied an entropy regularizer for the softmax attention as described in \newcite{kurach2015neural}.} The joint representation is the sum of the output vectors weighted by this attention (eqn.~\ref{eq:memnet_answer}).
\begin{align}
  I(x_i,W) &= \textstyle\sum_{j} Wx_{ij} \label{eq:memnet_encoder} \\
  \mathbf{u} &= I(p, U) \label{eq:memnet_question} \\
  \mathbf{m}_i &= I(x_i, M) \label{eq:memnet_memory} \\
  \mathbf{c}_i &= I(x_i, C) \label{eq:memnet_output} \\
  \mathbf{p}_i &= \softmax(\mathbf{q}^{\intercal}\mathbf{m}_i) \label{eq:memnet_attention} \\
  \mathbf{y} &= \mathbf{u} + \textstyle\sum_i \mathbf{p}_i \mathbf{c}_i \label{eq:memnet_answer}
\end{align}

\subsection{Answer Extraction}\label{sec:rnn_labeler}

Relational properties involve mappings between arbitrary entities (e.g., \texttt{date of birth}, \texttt{mother}, and \texttt{author}) and thus are less amenable to document classification. 
For these, approaches from information extraction (especially relation extraction) are much more appropriate. 
In general, these methods seek to identify a word or phrase in the text that stands in a particular relation to a (possibly implicit) subject.
Section \ref{sec:related} contains a discussion of prior work applying NLP techniques involving entity recognition and syntactic parsing to this problem.

\begin{figure}
\includegraphics[scale=0.40]{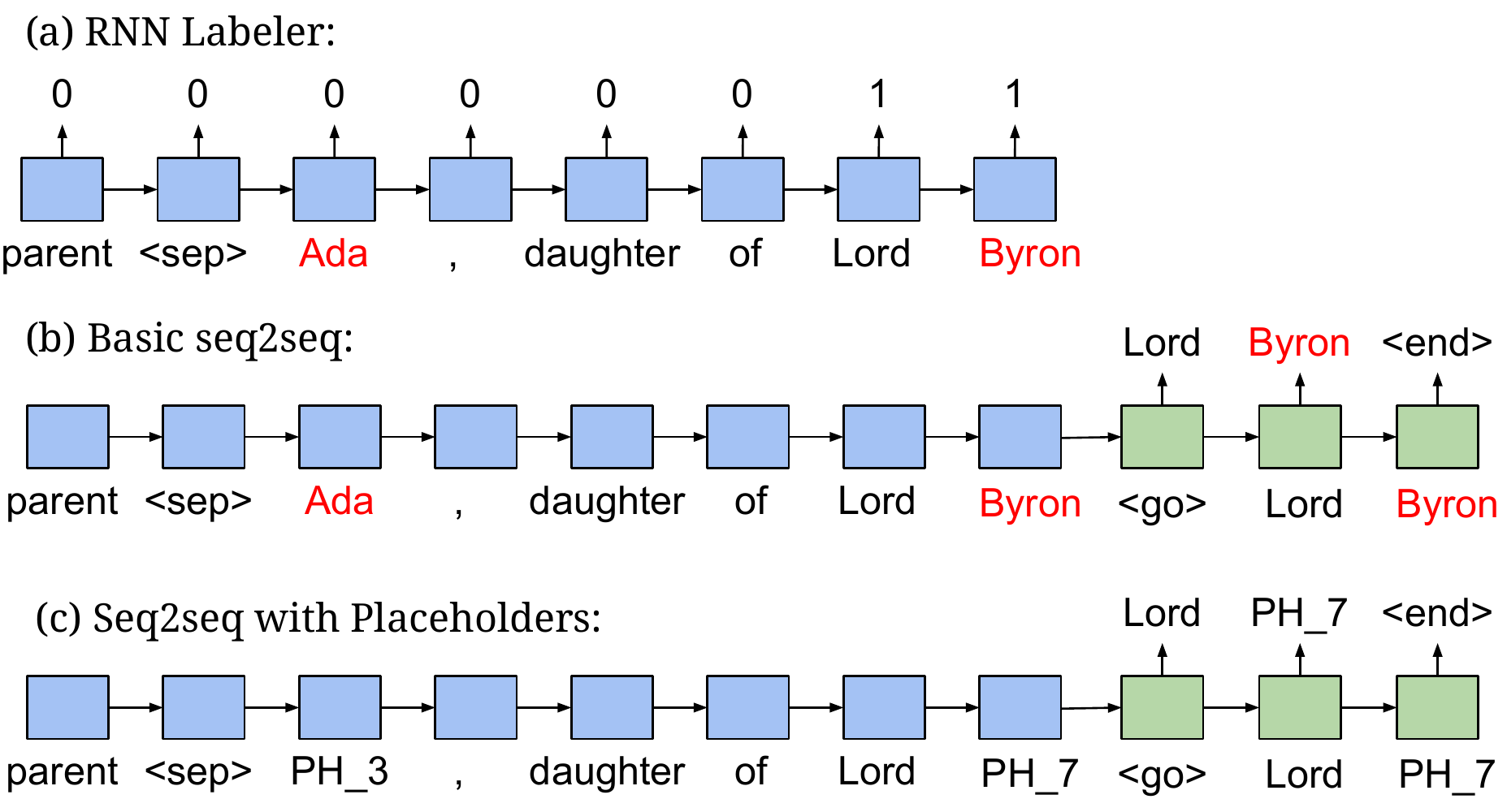}
\caption{Illustration of RNN models. Blocks with same color share parameters. Red words are out of vocabulary and all share a common embedding.}\label{fig:rnn_models}
\end{figure}

RNNs provide a natural fit for extraction, as they can predict a value at every position in a sequence, conditioned on the entire previous sequence. The most straightforward application to \textsc{WikiReading} is to predict the probability that a word at a given location is part of an answer. We test this approach using an RNN that operates on the sequence of words. At each time step, we use a sigmoid activation for estimating whether the current word is part of the answer or not. We refer to this model as the \emph{RNN Labeler} and present it graphically in Figure~\ref{fig:rnn_models}a. 

For training, we label all locations where any answer appears in the document with a $1$, and other positions with a $0$ (similar to distant supervision \cite{Mintz:2009:DSR:1690219.1690287}). For multi-word answers, the word sequences in the document and answer must fully match\footnote{Dates were matched semantically to increase recall.}. Instances where no answer appears in the document are discarded for training. The cost function is the average cross-entropy for the outputs across the sequence. When performing inference on the test set, sequences of consecutive locations scoring above a threshold are chunked together as a single answer, and the top-scoring answer is recorded for submission.\footnote{We chose an arbitrary threshold of 0.5 for chunking. The score of each chunk is obtained from the harmonic mean of the predicted probabilities of its elements.}

\subsection{Sequence to Sequence}

Recently, sequence to sequence learning (or \emph{seq2seq}) has shown promise for natural language tasks, especially machine translation \cite{cho-al-emnlp14}. 
These models combine two RNNs: an \emph{encoder}, which transforms the input sequence into a vector representation, and a \emph{decoder}, which converts the encoder vector into a sequence of output tokens, one token at a time.
This makes them capable, in principle, of approximating any function mapping sequential inputs to sequential outputs. 
%In this section, we explore the application of these models to \textsc{WikiReading}.
Importantly, they are the first model we consider that can perform any combination of answer classification and extraction.

\subsubsection{Basic seq2seq}\label{sec:basic_seq2seq}

This model resembles LSTM Reader augmented with a second RNN to decode the answer as a sequence of words. The embedding matrix is shared across the two RNNs but their state to state transition matrices are different (Figure~\ref{fig:rnn_models}b). This method extends the set of possible answers to any sequence of words from the document vocabulary. 

\subsubsection{Placeholder seq2seq}\label{sec:ph_sec2sec}

While Basic seq2seq already expands the expressiveness of LSTM Reader, it still has a limited vocabulary and thus is unable to generate some answers.
As mentioned in Section \ref{sec:rnn_labeler}, RNN Labeler can extract any sequence of words present in the document, even if some are OOV.
%, but is still unable to infer any answer that is not textually in the document.
%This suggests that extending seq2seq with the ability to copy OOV words from the document sequence would be beneficial. 
We extend the basic seq2seq model to handle OOV words by adding \emph{placeholders} to our vocabulary, increasing the vocabulary size from $\Nword$ to $\Nword + \Ndoc$.
Then, when an OOV word occurs in the document, it is replaced at random (without replacement). by one of these placeholders. 
We also replace the corresponding OOV words in the target output sequence by the same placeholder,\footnote{The same OOV word may occur several times in the document. Our simplified approach will attribute a different placeholder for each of these and will use the first occurrence for the target answer.} as shown in Figure \ref{fig:rnn_models}c. \newcite{luong-EtAl:2015:ACL-IJCNLP} developed a similar procedure for dealing with rare words in machine translation, copying their locations into the output sequence for further processing.

This makes the input and output sequences a mixture of known words and placeholders, and allows the model to produce any answer the RNN Labeler can produce, in addition to the ones that the basic seq2seq model could already produce. 
This approach is comparable to \emph{entity anonymization} used in \newcite{DBLP:journals/corr/HermannKGEKSB15}, which replaces named entities with random ids, but simpler because we use word-level placeholders without entity recognition.

\subsubsection{Basic Character seq2seq}\label{sec:char_seq2seq}

Another way of handling rare words is to process the input and output text as sequences of characters or bytes.
RNNs have shown some promise working with character-level input, including state-of-the-art performance on a Wikipedia text classification benchmark \cite{dai2015semi}.
A model that outputs answers character by character can in principle generate any of the answers in the test set, a major advantage for \textsc{WikiReading}.

\begin{figure}
 \begin{center}
 \includegraphics[width=0.45\textwidth]{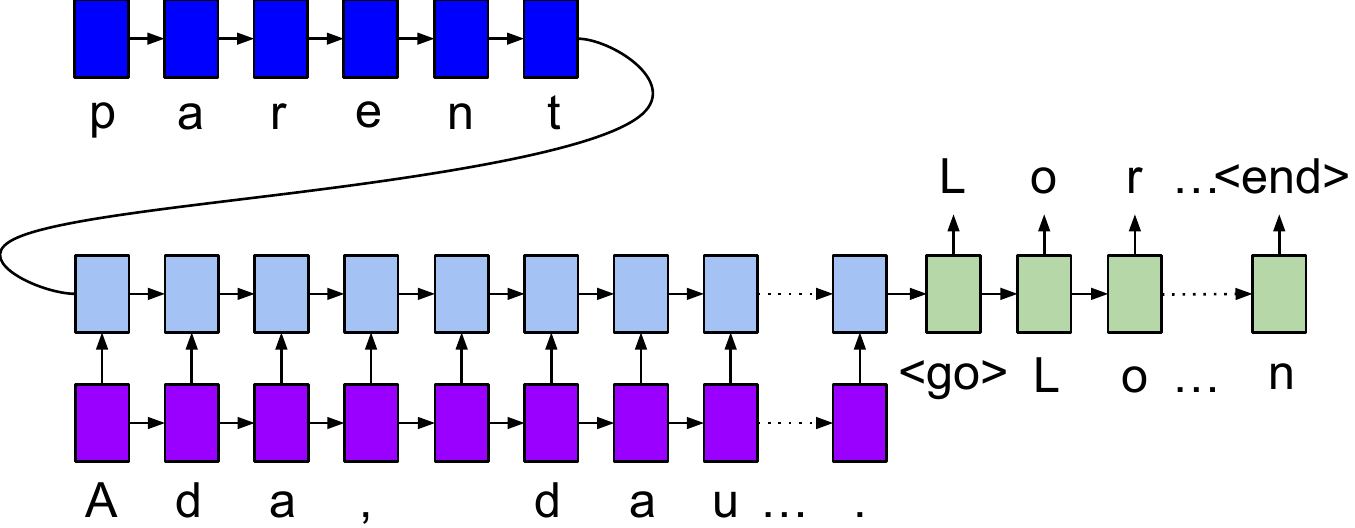} 
  \caption{Character seq2seq model. Blocks with the same color share parameters. The same example as in Figure~\ref{fig:rnn_models} is fed character by character. %More details on this model can be found in Section~\ref{sec:char_seq2seq}.
  } \label{fig:char-seq2seq2}
 \end{center} 
\end{figure}

This model, shown in Figure~\ref{fig:char-seq2seq2}, operates only on sequences of mixed-case characters.
The property encoder RNN transforms the property, as a character sequence, into a fixed-length vector. 
This property encoding becomes the initial hidden state for the second layer of a two-layer document encoder RNN, which reads the document, again, character by character. 
Finally, the answer decoder RNN uses the final state of the previous RNN to decode the character sequence for the answer.

\subsubsection{Character seq2seq with Pretraining}

Unfortunately, at the character level the length of all sequences (documents, properties, and answers) is greatly increased. 
This adds more sequential steps to the RNN, requiring gradients to propagate further, and increasing the chance of an error during decoding. 
To address this issue in a classification context, \newcite{dai2015semi} showed that initializing an LSTM classifier with weights from a language model (LM) improved its accuracy. 
Inspired by this result, we apply this principle to the character seq2seq model with a two-phase training process: In the first phase, we train a character-level LM on the input character sequences from the \textsc{WikiReading} training set (no new data is introduced). 
In the second phase, the weights from this LM are used to initialize the first layer of the encoder and the decoder (purple and green blocks in Figure~\ref{fig:char-seq2seq2}).
After initialization, training proceeds as in the basic character seq2seq model.

\newcommand{\grad}{\mathbf{g}}

\section{Experiments}\label{sec:experiments}

\begin{table*}[t]
\small 
\begin{center}
\begin{tabular}{|l|c|c|c|c|c|r|}
\hline
{\bf Method} & {\bf Mean F1} & {\bf Bound} & {\bf Categorical} & {\bf Relational} & {\bf Date} & {\bf Params} \\
\hline
\multicolumn{7}{|l|}{{\bf Answer Classifier}}\\
\hline
Sparse BoW Baseline & 0.438 & \multirow{6}{*}{0.831} & 0.725 & 0.063 & 0.004 & 500.5M \\  % FINAL
Averaged Embeddings & 0.583 && 0.849 & 0.234 & 0.080 & 120M \\  % FINAL
Paragraph Vector & 0.552 && 0.787 & 0.227 & 0.033 & 30M \\
LSTM Reader & 0.680 && 0.880 & 0.421 & 0.311 & 45M \\  % FINAL
Attentive Reader & 0.693 && {\bf 0.886} & 0.441 & 0.337 & 56M \\  % FINAL
Memory Network & 0.612 && 0.861 & 0.288 & 0.055 & 90.1M \\  % FINAL
\hline
\multicolumn{7}{|l|}{{\bf Answer Extraction}}\\
\hline
RNN Labeler & 0.357 & 0.471 & 0.240 & 0.536 & 0.626 & 41M \\
\hline
\multicolumn{7}{|l|}{{\bf Sequence to Sequence}}\\
\hline
Basic seq2seq & 0.708 & 0.925 & 0.844 & 0.530 & {\bf 0.738} & 32M \\
Placeholder seq2seq & 0.{\bf 718} & 0.948 & 0.835 &  {\bf 0.565} & 0.730 & 32M \\
Character seq2seq & 0.677 & {\bf 0.963} & 0.841 & 0.462 & 0.731 & 4.1M \\  % FINAL
Character seq2seq (LM) & 0.699 & {\bf 0.963} & 0.851 &  0.501 &  0.733 & 4.1M \\
\hline
\end{tabular}
\end{center}
\caption{
Results for all methods described in Section \ref{sec:methods} on the test set. 
{\it F1} is the Mean F1 score described in \ref{sec:experiments}. 
{\it Bound} is the upper bound on Mean F1 imposed by constraints in the method (see text for details). 
The remaining columns provide score breakdowns by property type and the number of model parameters.
}\label{tab:results}
\end{table*}

%% Gaggle Links
% * Basic seq2seq: https://gaggle.corp.google.com/submission/944
% * Placeholder seq2seq: https://gaggle.corp.google.com/submission/869
% * Char seq2seq: https://gaggle.corp.google.com/submission/948
% * Char seq2seq (LM): https://gaggle.corp.google.com/submission/947

%To ensure that no answers for the test documents are seen during training, we divide the \emph{Wikipedia articles} into training, validation, and test sets following a 70/10/20 ratio. All instances for a given article are added to its respective set, yielding a training set of 11.29M instances, validation set of 1.60M instances, and test set of 3.12M instances.

We evaluated all methods from Section~\ref{sec:methods} on the full test set with a single scoring framework. An answer is correct when there is an exact string match between the predicted answer and the gold answer. However, as describe in Section~\ref{sec:dataset}, some answers are composed from a set of values (e.g. third example in Table~\ref{tab:examples}). To handle this, we define the \emph{Mean F1} score as follows: For each instance, we compute the F1-score (harmonic mean of precision and recall) as a measure of the degree of overlap between the predicted answer set and the gold set for a given instance. The resulting per-instance F1 scores are then averaged to produce a single dataset-level score. This allows a method to obtain partial credit for an instance when it answers with at least one value from the golden set. In this paper, we only consider methods for answering with a single value, and most answers in the dataset are also composed of a single value, so this Mean F1 metric is closely related to accuracy. More precisely, a method using a single value as answer is bounded by a Mean F1 of 0.963.

%o assign some credit for partially correct answer sets, 
%For the 90\% of the instances in the dataset that have only one element in the answer set, these two scoring systems are identical.

% Table created from https://gaggle.corp.google.com/submission/diff/3825?oid=734&oid=816&oid=746&oid=944&oid=869&oid=948&oid=947

\subsection{Training Details}\label{sec:training}

We implemented all models in a single framework based on TensorFlow \cite{abaditensorflow} with shared pre-processing and comparable hyperparameters whenever possible. 
All documents are truncated to the first $300$ words except for Character seq2seq, which uses $400$ characters. 
The embedding matrix used to encode words in the document uses $\din=300$ dimensions for the $\Nword=100,000$ most frequent words. 
Similarly, answer classification over the $\Nans=50,000$ most frequent answers is performed using an answer representation of size $\dout=300$.\footnote{For models like Averaged Embedding and Paragraph Vector, the concatenation imposes a greater $\dout$.} 
The first $10$ words of the properties are embedded using the document embedding matrix. 
%Memory networks use 3 layers of reasoning and 
%TODO update to reflect new memory network section
Following \newcite{cho-al-emnlp14}, RNNs in seq2seq models use a GRU cell with a hidden state size of $1024$. More details on parameters are reported in Table~\ref{tab:params}.

\begin{table}[h]
\begin{centering}
\small
\begin{tabular}{|m{1.3cm}|m{0.9cm}|m{1cm}|m{1.2cm}|m{1cm}|}
\hline
{\bf Method} & {\bf Emb. Dims} & {\bf Doc. Length} & {\bf Property Length} & {\bf Doc. Vocab. Size}\\
\hline
Sparse BoW Baseline & N/A & 300 words & 10 words & 50K words\\\hline
Paragraph Vector & N/A & N/A & 10 words & N/A \\\hline
Character seq2seq & 30 & 400 chars & 20 chars & 76 chars \\\hline
All others & 300 & 300 words & 10 words & 100K words \\\hline
\end{tabular}
\caption{Structural model parameters. Note that the Paragraph Vector method uses the output from a separate, unsupervised model as a document encoding, which is not counted in these parameters. }\label{tab:params}
\end{centering}
\end{table}

% OLD gradient clipping citation: DBLP:conf/icml/PascanuMB13
Optimization was performed with the Adam stochastic optimizer\footnote{Using $\beta_1=0.9$, $\beta_2=0.999$ and $\epsilon=10^{-8}$.} \cite{kingma2015method} over mini-batches of $128$ samples. 
Gradient clipping \footnote{When the norm of gradient $\grad$ exceeds a threshold $C$, it is scaled down i.e. $\grad \leftarrow \grad \cdot \min \left( 1, \frac{C}{||\grad||} \right)$.} \cite{DBLP:journals/corr/Graves13} is used to prevent instability in training RNNs. 
%To determine effective values for the learning rate and the gradient clip threshold, 
We performed a search over 50 randomly-sampled hyperparameter configurations for the learning rate and gradient clip threshold, selecting the one with the highest Mean F1 on the validation set.
Learning rate and clipping threshold are sampled uniformly, on a logarithmic scale, over the range $[10^{-5},10^{-2}]$ and $[10^{-3},10^{1}]$ respectively.

%\begin{table}
%\begin{center}
%\small
%\begin{tabular}{|l|l|}
%\hline
%{\bf Parameter} & {\bf Value} \\\hline
%Learning rate (LR) & 0.00001 to 0.01 \\\hline
%Gradient clip (GC) & 0.001 to 1.0 \\\hline
%(LR,GC) values searched & 50 \\\hline
%Number of epochs & 4 \\\hline
%Hidden state size & 1024 \\\hline
%\end{tabular}
%\end{center}
%\caption{Shared hyperparameter search values.}\label{tab:training-params}
%\end{table}

\subsection{Results and Discussion}\label{sec:results}

\begin{table*}[t]
\small 
\begin{center}
\begin{tabular}{|m{1.41cm}|m{1.0cm}|m{1.4cm}|m{1.25cm}|m{1.4cm}|m{1.2cm}|m{1.6cm}|m{1.4cm}|m{1.4cm}|}
\hline
& & \multicolumn{7}{c|}{{\bf Mean F1}} \\\cline{3-9}
{\bf Property} & {\bf Test Instances} & {\bf Averaged Embeddings} & {\bf Attentive Reader} & {\bf Memory Network} & {\bf Basic seq2seq} & {\bf Placeholder seq2seq} & {\bf Character seq2seq} & {\bf Character seq2seq (LM)} \\
\hline
\multicolumn{9}{|l|}{{\bf Categorical Properties}}\\
\hline
instance of & 191157 & 0.8545 & {\bf 0.8978} & 0.8720 & 0.8877 & 0.8775 & 0.8548 & 0.8659 \\\hline
sex or gender & 66875 & 0.9917 & 0.9966 & 0.9936 & {\bf 0.9968} & 0.9952 & 0.9943 & 0.9941 \\\hline
genre & 8117 &	0.5320 &	{\bf 0.6225 } &	0.5625 &	0.5511 &	0.5260 &	0.5096 & 	0.5283 \\\hline
instrument & 890 &	0.7621 & {\bf 0.8415 } & 0.7886 &	0.8377 &	0.8172 & 	0.7529 & 	0.7832 \\\hline
\multicolumn{9}{|l|}{{\bf Relational Properties}}\\
\hline
given name & 54624 & 0.4973 & 0.8486 & 0.7206 & 0.8669 & {\bf 0.8868} & 0.8606 & 0.8729 \\\hline
located in & 34250 &	0.4140 & 0.6195 & 0.4832 & 0.5484 & {\bf 0.6978} & 0.5496 & 0.6365 \\\hline
% place of birth & 109546 & 0.2550 & 0.5043 & 0.3513 & 0.4892 & {\bf 0.5217} & 0.3632 & 0.4061 \\\hline
parent taxon & 15494 & 0.1990 & 0.3467 & 0.2077 & 0.2044 & {\bf 0.7997} & 0.4979 & 0.5748 \\\hline
author & 2356 & 0.0309 & 0.2088 & 0.1050 & 0.6094 &	{\bf 0.6572} & 0.1403 & 0.3748 \\\hline
\multicolumn{9}{|l|}{{\bf Date Properties}}\\
\hline
date of birth & 56221 & 0.0626 & 0.3677 & 0.0016 & {\bf 0.8306} & 0.8259 & 0.8294 & 0.8303 \\\hline
date of death & 26213 & 0.0417 & 0.2949 & 0.0506 & {\bf 0.7974} & 0.7874 & 0.7897 & 0.7924 \\\hline
publication date & 7680 & 0.3909 & 0.5549 & 0.4851 & {\bf 0.5988} & 0.5902 & 0.5903 & 0.5943 \\\hline
date of official opening & 269 & 0.1510 & 0.3047 & 0.1725 & {\bf 0.3333} & 0.3012 & 0.1457 & 0.1635 \\\hline
\end{tabular}
\end{center}
\caption{Property-level Mean F1 scores on the test set for selected methods and properties. For each property type, the two most frequent properties are shown followed by two less frequent properties to illustrate long-tail behavior.}\label{tab:results-per-relation}
\end{table*}

%TODO from joberant:
%Section 4: 
%* If I understand correctly the best result is obtained from reading the first 300 characters, i.e. ~ 100 words? wow. 
%
%* I think I wrote that but the character result is pretty exciting I think and should perhaps appear more prominently in the introduction and also later with something like "interestingly, a character based model achieves best performance" and other similar things. 
%
%* I suspect someone might ask you what happens when you ensemble the attentive reader with the rnn labeler and so perhaps it's good to pre-empt and present that - I remember you had that in the past. 

Results for all models on the held-out set of test instances are presented in Table \ref{tab:results}.
In addition to the overall Mean F1 scores, the model families differ significantly in Mean F1 upper bound, and their relative performance on the relational and categorical properties defined in Section \ref{sec:properties}.
We also report scores for properties containing dates, a subset of relational properties, as a separate column since they have a distinct format and organization.
For examples of model performance on individual properties, see Table \ref{tab:results-per-relation}.
%To provide more context for the Mean F1 scores, we also computed the maximum possible score achievable by each model, based on inherent constraints of the approach. 
%For example, the answer classification models from Section \ref{sec:answer-classification} can only choose from a fixed set of answers, and so automatically fail every instance with an answer outside this set. 
%In addition, we also partition the test instances into those with categorical and relational properties (as defined in Section \ref{sec:properties}), and report results on each subset. 
%Finally, we provide the number of parameters of the model variant that achieved the score shown in the table (for details of model training, see Section \ref{sec:training} below).

\begin{figure}
 \begin{center}
 \includegraphics[width=0.45\textwidth]{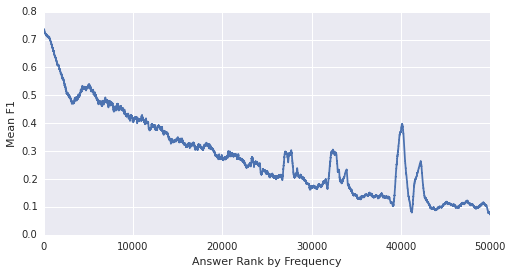}
  \caption{Per-answer Mean F1 scores for Attentive Reader (moving average of 1000), illustrating the decline in prediction quality as the number of training examples per answer decreases.}\label{fig:ans_freq}
 \end{center} 
\end{figure}

As expected, all classifier models perform well for categorical properties, with more sophisticated classifiers generally outperforming simpler ones.
The difference in precision reading ability between models that use broad document statistics, like Averaged Embeddings and Paragraph Vectors, and the RNN-based classifiers is revealed in the scores for relational and especially date properties.
As shown in Table \ref{tab:results-per-relation}, this difference is magnified in situations that are more difficult for a classifier, such as relational properties or properties with fewer training examples, where Attentive Reader outperforms Averaged Embeddings by a wide margin.
This model family also has a high upper bound, as perfect classification across the $50,000$ most frequent answers would yield a Mean F1 of 0.831.
However, none of them approaches this limit.
Part of the reason is that their accuracy for a given answer decreases quickly as the frequency of the answer in the training set decreases, as illustrated in Figure \ref{fig:ans_freq}.
As these models have to learn a separate weight vector for each answer as part of the softmax layer (see Section \ref{sec:answer-classification}), this may suggest that they fail to generalize across answers effectively and thus require significant number of training examples per answer.

The only answer extraction model evaluated, RNN Labeler, shows a complementary set of strengths, performing better on relational properties than categorical ones. 
While the Mean F1 upper bound for this model is just 0.434 because it can only produce answers that are present verbatim in the document text, it manages to achieve most of this potential.
The improvement on date properties over the classifier models demonstrates its ability to identify answers that are typically present in the document.
We suspect that answer extraction may be simpler than answer classification because the model can learn robust patterns that indicate a location without needing to learn about each answer, as the classifier models must.

The sequence to sequence models show a greater degree of balance between relational and categorical properties, reaching performance consistent with classifiers on the categorical questions and with RNN Labeler on relational questions.
Placeholder seq2seq can in principle produce any answer that RNN Labeler can, and the performance on relational properties is indeed similar.
As shown in Table \ref{tab:results-per-relation}, Placeholder seq2seq performs especially well for properties where the answer typically contains rare words such as the name of a place or person.
When the set of possible answer tokens is more constrained, such as in categorical or date properties, the Basic seq2seq often performs slightly better.
Character seq2seq has the highest upper bound, limited to 0.963 only because it cannot produce an answer set with multiple elements.
LM pretraining consistently improves the performance of the Character seq2seq model, especially for relational properties as shown in Table \ref{tab:results-per-relation}.
The performance of the Character seq2seq, especially with LM pretraining, is a surprising result:
It performs comparably to the word-level seq2seq models even though it must copy long character strings when doing extraction and has access to a smaller portion of the document. 
We found the character based models to be particularly sensitive to hyperparameters. 
However, using a pretrained language model reduced this issue and significantly accelerated training while improving the final score. 
We believe that further research on pretraining for character based models could improve this result. 

\section{Related Work}\label{sec:related}

The goal of automatically extracting structured information from unstructured Wikipedia text was first advanced by \newcite{wu2007autonomously}.
As Wikidata did not exist at that time, the authors relied on the structured \emph{infoboxes} included in some Wikipedia articles for a relational representation of Wikipedia content. 
Wikidata is a cleaner data source, as the infobox data contains many slight variations in schema related to page formatting.
Partially to get around this issue, the authors restrict their prediction model Kylin to 4 specific infobox classes, and only common attributes within each class. 

A substantial body of work in relation extraction (RE) follows the distant supervision paradigm \cite{Craven:1999:CBK:645634.663209}, where sentences containing both arguments of a knowledge base (KB) triple are assumed to express the triple's relation.
Broadly, these models use these distant labels to identify syntactic features relating the subject and object entities in text that are indicative of the relation.
\newcite{Mintz:2009:DSR:1690219.1690287} apply distant supervision to extracting Freebase triples \cite{Bollacker:2008:FCC:1376616.1376746} from Wikipedia text, analogous to the relational part of \textsc{WikiReading}. 
Extensions to distant supervision include explicitly modelling whether the relation is actually expressed in the sentence \cite{riedel2010modeling}, and jointly reasoning over larger sets of sentences and relations \cite{surdeanu2012multi}.
Recently, \newcite{rocktaschel2015injecting} developed methods for reducing the number of distant supervision examples required by sharing information between relations.

\section{Conclusion}

We have demonstrated the complexity of the \textsc{WikiReading} task and its suitability as a benchmark  to guide future development of DNN models for natural language understanding.
After comparing a diverse array of models spanning classification and extraction, we conclude that end-to-end sequence to sequence models are the most promising.
These models simultaneously learned to classify documents and copy arbitrary strings from them. 
In light of this finding, we suggest some focus areas for future research.

Our character-level model improved substantially after language model pretraining, suggesting that further training optimizations may yield continued gains.
Document length poses a problem for RNN-based models, which might be addressed with convolutional neural networks that are easier to parallelize.
Finally, we note that these models are not intrinsically limited to English, as they rely on little or no pre-processing with traditional NLP systems. 
This means that they should generalize effectively to other languages, which could be demonstrated by a multilingual version of \textsc{WikiReading}.

\section*{Acknowledgments}

We thank Jonathan Berant for many helpful comments on early drafts of the paper, and Catherine Finegan-Dollak for an early implementation of RNN Labeler.

\bibliography{acl2016}
\bibliographystyle{acl2016}

\end{document}